\pdfoutput=1

\documentclass[letterpaper, 10pt, conference]{ieeeconf}   %

\IEEEoverridecommandlockouts                              %

\let\labelindent\relax                                    %

\usepackage{silence}      %
\usepackage{graphicx}     %
\usepackage{amsmath}      %
\usepackage{amssymb}      %
\usepackage{stmaryrd}     %
\usepackage{amsfonts}     %
\usepackage{enumitem}     %
\usepackage{multirow}     %
\usepackage{siunitx}      %
\usepackage{tensor}       %
\usepackage{url}          %
\usepackage{calc}         %
\usepackage{xcolor}       %
\usepackage{xspace}       %
\usepackage{printlen}     %
\usepackage[T1]{fontenc}  %
\usepackage{hyperref}     %

\hypersetup{hidelinks, bookmarksdepth=3, bookmarksopen, bookmarksnumbered=false, pdfauthor=Philipp Allgeuer}

\makeatletter
\let\MYcaption\@makecaption
\makeatother
\usepackage[font=footnotesize]{subcaption}
\makeatletter
\let\@makecaption\MYcaption
\makeatother

\makeatletter
\def\subsubsection{\@startsection{subsubsection}{3}{\parindent}{0ex plus 0.1ex minus 0.1ex}{0ex}{\normalfont\normalsize\bfseries}}
\makeatother

\makeatletter
\newcommand*\bigcdot{\mathpalette\bigcdot@{.5}}
\newcommand*\bigcdot@[2]{\mathbin{\vcenter{\hbox{\scalebox{#2}{$\m@th#1\bullet$}}}}}
\makeatother

\graphicspath{{images/}}
\DeclareGraphicsExtensions{.pdf,.png,.jpg,.jpeg}

\setlist[itemize]{leftmargin=\parindent+3pt,labelindent=\parindent}
\setlist[enumerate]{leftmargin=*}

\DeclareMathOperator{\sgn}{sgn}

\providecommand{\abs}[1]{\left\lvert#1\right\rvert}
\providecommand{\norm}[1]{\left\lVert#1\right\rVert}

\newcommand{\vect}[1]{\mathbf{#1}} %
\newcommand{\yawof}[1]{\Psi\bigl(#1\bigr)} %
\newcommand{\fr}[1]{\{#1\}\xspace} %
\newcommand{\conj}{^{*}} %
\renewcommand{\P}{\mathbb{P}} %
\newcommand{\R}{\mathbb{R}} %
\newcommand{\T}{\mathbb{T}} %
\newcommand{\degreem}{^{\circ}} %
\newcommand{\seclabel}[1]{\label{sec:#1}}

\newcommand{\figlabel}[1]{\label{fig:#1}}
\newcommand{\tablabel}[1]{\label{tab:#1}}
\newcommand{\eqnlabel}[1]{\label{eqn:#1}}

\newcommand{\secref}[1]{Section~\ref{sec:#1}\xspace}

\newcommand{\figref}[1]{Fig.~\ref{fig:#1}\xspace}
\newcommand{\tabref}[1]{Table~\ref{tab:#1}\xspace}
\newcommand{\eqnref}[1]{(\ref{eqn:#1})\xspace}

\newcommand{\code}[1]{\protect\path{#1}} %

\newcommand{\noptwo}{NimbRo\protect\nobreakdash-OP2\xspace}

\newcommand{\iguhop}{igus\textsuperscript{\tiny\circledR}$\!$ Humanoid Open Platform\xspace}

\newcommand{\cpp}{C\texttt{\nolinebreak\hspace{-.05em}+\nolinebreak\hspace{-.05em}+}\xspace}

\newcommand{\degree}{$\degreem$\xspace}

\title{\LARGE \bf Bipedal Walking with Corrective Actions in the Tilt Phase Space}

\author{Philipp Allgeuer and Sven Behnke%
\thanks{All authors are with the Autonomous Intelligent Systems (AIS) Group, Computer Science Institute VI,
        University of Bonn, Germany. Email: {\tt\small pallgeuer@ais.uni-bonn.de}. This work was partially
        funded by grant BE 2556/13 of the German Research Foundation (DFG).}}

\usepackage{eso-pic}

\AtBeginDocument{\AddToShipoutPictureFG*{\AtTextUpperLeft{\put(0,\LenToUnit{9pt}){\parbox{\textwidth}{\centering\bfseries
International Conference on Humanoid Robots (Humanoids), Beijing, China, 2018
}}}}}
\begin{document}

\bstctlcite{IEEEexample:BSTcontrol}

\maketitle
\thispagestyle{empty}
\pagestyle{empty}

\begin{abstract}
Many methods exist for a bipedal robot to keep its balance while walking. In 
addition to step size and timing, other strategies are possible that influence 
the stability of the robot without interfering with the target direction and 
speed of locomotion. This paper introduces a multifaceted feedback controller 
that uses numerous different feedback mechanisms, collectively termed corrective 
actions, to stabilise a core keypoint-based gait. The feedback controller is 
experimentally effective, yet free of any physical model of the robot, very 
computationally inexpensive, and requires only a single 6-axis IMU sensor. Due 
to these low requirements, the approach is deemed to be highly portable between 
robots, and was specifically also designed to target lower cost robots that have 
suboptimal sensing, actuation and computational resources. The IMU data is used 
to estimate the yaw-independent tilt orientation of the robot, expressed in the 
so-called tilt phase space, and is the source of all feedback provided by the 
controller. Experimental validation is performed in simulation as well as on 
real robot hardware.
\end{abstract}

\section{Introduction}
\seclabel{introduction}

Many feedback strategies exist by which a robot can seek to maintain its balance 
while walking bipedally. Often considered is the online adjustment of step size 
and timing, e.g.\ \cite{Kryczka2015}. While these are quite effective strategies 
if done right, numerous other forms of feedback beyond just ankle torque, such 
as for example arm motions and swing leg trajectory adjustments, can also be 
employed to significantly increase the stability of the robot, especially in a 
broader spectrum of walking situations. For instance, step size feedback cannot 
help when a robot is about to tip over the outside of one of its feet, or 
effectively correct for systematic biases in the robot. It also has little 
effect until the next step is actually taken, meaning that there is an inherent 
dead time until disturbances can be counteracted. Furthermore, changing the 
target step size modifies the footholds, and thus directly leads to the 
non-realisation of footstep plans. As such, step size feedback is envisioned as 
a valuable tool for gait stabilisation, but one that ideally only activates for 
large disturbances, when there really is no other option. The corrective actions 
presented in this paper (see \figref{teaser}) aim to address all of these 
issues, in addition to solving the more general problem of how to achieve 
balanced push-resistant walking, with minimal changes to the walking intent of 
the robot.

In the interest of reducing the required tuning effort and making the gait 
applicable to low-cost robots with cheap sensors and actuators, the use of 
physical models in the feedback path is avoided. Physical models usually require 
extensive model identification and tuning to sufficiently resemble the behaviour 
of a robot, and even then, cheap actuators lead to significant nonlinearities 
that can often cause such models to have poor results or fail. Physical models 
are also frequently quite sensitive to small changes in the robot, leading to 
frequent retuning being necessary. The implementation difficulty and cost of 
good sensors also limits the type and accuracy of sensors that can be 
incorporated into a humanoid robot. So to facilitate the greatest possible 
portability of the developed gait between robots of different builds and 
proportions---a design decision that is supported by the chosen model-free 
nature of the gait---only the presence of a 6-axis IMU sensor is assumed. Apart 
from that, no additional sensors, joint positions, robot masses or inertias are 
assumed at all for the feedback controller. The only further `assumptions' that 
are made are trivial notions, such as for example that tilting the foot in one 
direction makes the robot tend to tilt in the other.

\begin{figure}[!t]
\parbox{\linewidth}{\centering\includegraphics[width=0.95\linewidth]{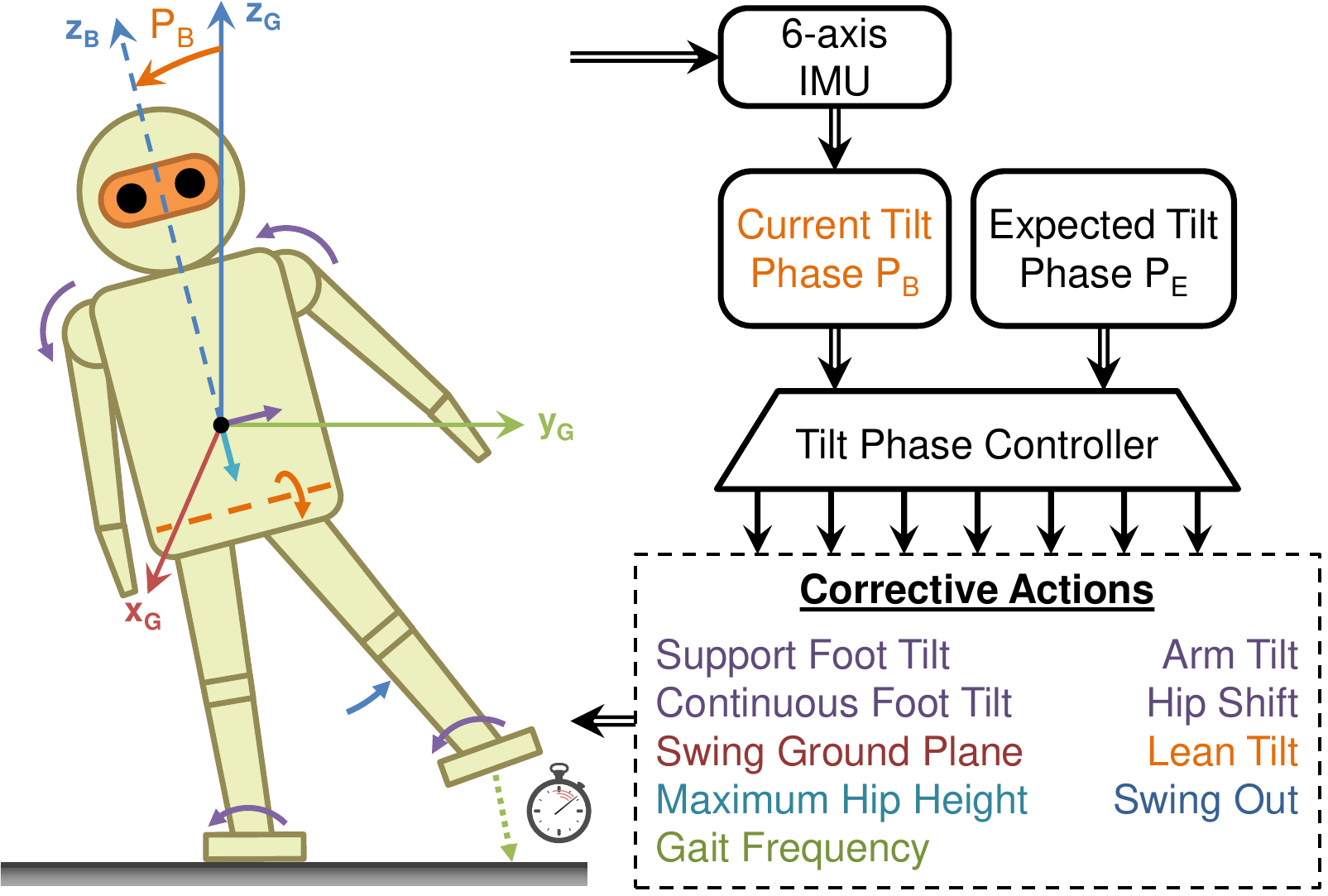}}
\caption{Overview of the tilt phase corrective action approach to walking.}
\figlabel{teaser}
\vspace{-3ex}
\end{figure}

The main contributions of this paper lie in the methods of calculation of the 
various feedback components, which are extensions of previous work 
\cite{Allgeuer2016a} only for the PD, leaning and timing components, and 
otherwise novel. Several corrective actions are also completely new, with the 
remaining ones being extended to a full 3D treatment, so as in particular not to 
treat the sagittal and lateral directions independently. This is aided by the 
novel use of the tilt phase space for truly axisymmetric orientation feedback. 
While the controller for the individual corrective actions is discussed in 
detail, how these actions are combined to form a single resulting joint 
trajectory is beyond the scope of this paper. The presented feedback controller 
has been released open source in \cpp \cite{iguhopSoftware}. Ultimately, this 
paper seeks to demonstrate that not overly complex feedback mechanisms with very 
limited information of the robot suffice to produce a very stable gait, capable 
of rejecting significant disturbances.

\section{Related Work}
\seclabel{related_work}

Many different approaches to the stabilisation of bipedal walking can be found 
in literature. Methods that are effective in actively rejecting large 
disturbances, yet do not use a physical model and only use a single inertial 
measurement unit (IMU) for feedback, are however infrequently encountered. A 
large proportion of walking methods, especially those for stiffer and higher 
quality robots, depend on zero-moment point (ZMP) based gait generation and/or 
feedback. Usually this involves optimising for a future centre of mass (CoM) 
trajectory based on certain targets and constraints, and then trying to execute 
this as closely as possible with a lower level controller. Examples of this 
include in the settings of model predictive control \cite{Wieber2006}, preview 
control \cite{Kajita2003} \cite{Urata2011}, and an indirect generalisation in 
the form of differential dynamic programming (DDP) \cite{Feng2013}. Tedrake et 
al. \cite{Tedrake2015} also proposed a closed-form solution to the full 
continuous-time ZMP LQR problem over large horizons, while Kajita et al. 
\cite{Kajita2017} use spatially quantised dynamics (SQD), as opposed to temporal 
quantisation, with a mix of reference ZMP trajectory generation and divergent 
component of motion (DCM) dynamics to generate a gait that was able to utilise 
fully stretched knees.

The main problem with implementing ZMP-based methods on low-cost robots is that 
such methods generally assume that actuator tracking is quite good, and that if 
theoretically stable motions are planned, then executing them will be 
essentially stable. ZMP reference tracking controllers aim to increase the 
margin of convergence and stability, but ZMP approaches frequently are just not 
designed to operate to the limits of stability, where for example a robot is 
completely tilted about one of the edges or even vertices of its feet, with 
significant energy and deviation from upright. ZMP tracking controllers also 
usually require direct sensing of the centre of pressure, which is usually not 
robustly available for limited sensing robots. The reliance on complex 
optimisation-based methods also means that computational requirements are high, 
as compared to the alternative of analytic computation.

The most closely related work to this paper is the work of the authors on direct 
fused angle feedback mechanisms \cite{Allgeuer2016a}. The main differences and 
improvements of this paper are explicitly outlined in the aforementioned list of 
contributions.

\section{The Tilt Phase Space}
\seclabel{tilt_phase_space}

One significant difference between \cite{Allgeuer2016a} and this paper is the 
full 3D treatment given to the corrective actions, made possible in part by 
the use of the \emph{tilt phase space} \cite{TiltPhase} instead of \emph{fused 
angles} \cite{Allgeuer2015}. While fused angles work very well for separate 
treatments of the sagittal and lateral planes, the tilt phase space has 
advantages for concurrent treatments, in particular in relation to magnitude 
axisymmetry \cite{TiltPhase}. This is important to ensure that feedback 
magnitudes are the same scale no matter what continuous direction the robot is 
tilted in. Furthermore, the tilt phase parameters share all of the critical 
advantages \cite{Allgeuer2018a} that fused angles have over lesser options, such 
as for example Euler angles, mainly due to the tight relationship between the 
two representations. Further advantages of the tilt phase space include that it 
can naturally represent and deal with tilt rotations of more than 180\degree, 
and that using it, tilt rotations can be unambiguously commutatively added. 
Both of these are useful features in feedback scenarios where rotation deviation 
feedback components are scaled by gains and need to be combined.

If $(\psi, \gamma, \alpha) \in \T$ are the \emph{tilt angles} parameters of a 
rotation, where $\psi$ is the fused yaw, $\gamma$ is the tilt axis angle and 
$\alpha$ is the tilt angle \cite{Allgeuer2015}, the corresponding \emph{3D tilt 
phase} is
\begin{equation}
P = (p_x, p_y, p_z) = (\alpha\cos\gamma, \alpha\sin\gamma, \psi) \in \R^3 \equiv \P^3. \eqnlabel{tiltphase3Ddefn}
\end{equation}
Omitting the yaw component, the \emph{2D tilt phase} representation of the 
resulting tilt rotation component is given by
\begin{equation}
\begin{gathered}
P = (p_x, p_y) \in \R^2 \equiv \P^2, \\
\begin{aligned}
\mspace{4mu} p_x &= \alpha\cos\gamma, &\mspace{77mu} p_y &= \alpha\sin\gamma.
\end{aligned}
\end{gathered}
\eqnlabel{tiltphase2Ddefn}
\end{equation}
Compare this to the following expressions for fused angles,
\begin{align}
\sin\phi &= \sin\alpha \cos\gamma, & \sin\theta &= \sin\alpha \sin\gamma. \eqnlabel{fused2Ddefn}
\end{align}
More details on all of the used rotation representations can be found in the 
respective papers \cite{TiltPhase} \cite{Allgeuer2015} and code 
releases.\footnote{\scriptsize \hspace{1pt}\cpp$\mspace{-6mu}/$Matlab: 
\url{https://github.com/AIS-Bonn/rot_conv_lib}\\
\url{https://github.com/AIS-Bonn/matlab_octave_rotations_lib}}

\begin{figure*}[!t]
\parbox{\linewidth}{\centering%
\includegraphics[height=4.9cm]{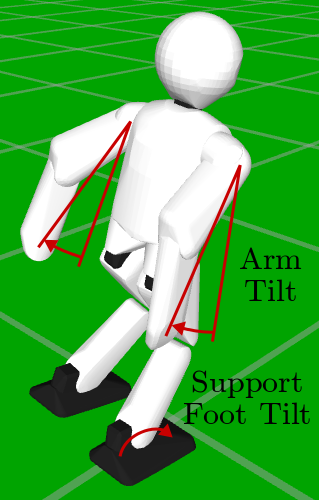}\hspace{1pt}%
\includegraphics[height=4.9cm]{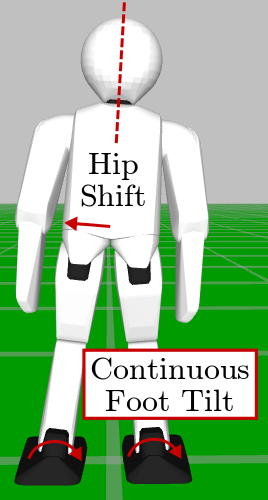}\hspace{1pt}%
\includegraphics[height=4.9cm]{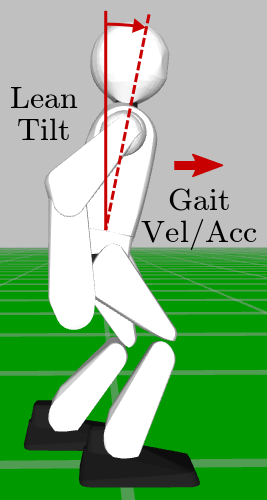}\hspace{1pt}%
\includegraphics[height=4.9cm]{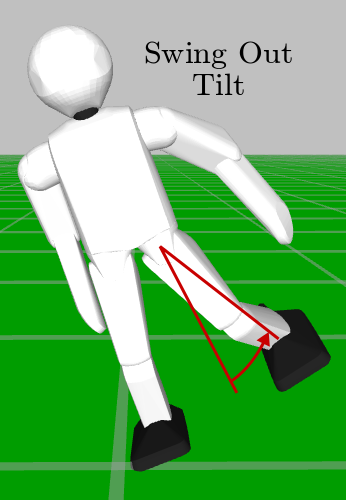}\hspace{1pt}%
\includegraphics[height=4.9cm]{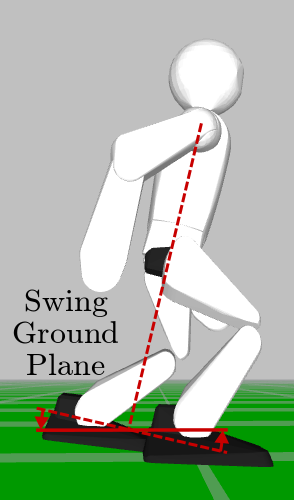}\hspace{1pt}%
\includegraphics[height=4.9cm]{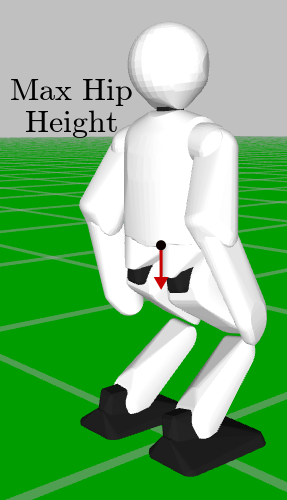}\hspace{1pt}}
\caption{Diagram of the various corrective actions implemented in the gait. In 
most cases, the images and annotations are gross 1D simplifications of the true 
corrective actions, for illustrative and explanatory purposes. The arrows 
indicate the effect of the corrective actions. The annotations in the swing 
ground plane image are trying to show that due to the forwards tilt of the 
robot, a normal step (lower dashed line) would have collided with the ground, 
while the adjusted step (solid line) avoids premature contact with the ground 
and executes the required step size despite the forwards tilt.}
\figlabel{corrective_actions}
\vspace{-2ex}
\end{figure*}
\section{Gait Architecture}
\seclabel{gait_architecture}

\subsection{Overview}
\seclabel{gait_overview}

The gait architecture consists of three layers, evaluated at \SI{100}{\hertz}, 
spanning from the generation of low level servo commands in the bottom layer, 
through to the generation of the required joint trajectories in the middle 
layer, and the calculation of feedback corrective action activation values in 
the top layer. The individual layers are discussed as follows.

\subsubsection{Actuator control scheme}
\seclabel{actuator_control_scheme}

In the lowest layer, the joint targets, joint efforts and support coefficients 
commanded by the middle layer are converted into raw actuator commands and 
stiffnesses that are communicated to the actuator hardware. This incorporates an 
optional feedforward torque component for improved tracking of the required 
trajectories. The actuator control scheme in use is thoroughly discussed in 
\cite{Allgeuer2016a}, and is useful to compensate for factors such as joint 
friction, battery voltage and the expected relative loadings of the legs---all 
nominally in a position-controlled setting.

\subsubsection{Keypoint gait generator}
\seclabel{keypoint_gait_generator}

The middle layer of the gait, referred to as the keypoint gait generator, is 
responsible for generating the required joint trajectories based on activation 
commands from the top layer. The implementation is something akin to an 
open-loop gait, albeit with the ability to activate a myriad of corrective 
actions, such as for example foot motions and hip shifts, which are ingrained 
into the construction of the gait. A full list is given in 
\secref{corrective_actions}. Note that the keypoint gait generator is totally 
different to the central pattern generated gaits used in previous publications. 
Also, the implementations of the corrective actions, unlike for example in 
\cite{Allgeuer2016a}, are given a full explicit 3D treatment, and are handled as 
proper rotations and transformations in 3D space, as opposed to independent 
contributions to joint angles based on separate lateral and sagittal 
contributions. Nevertheless, even with no activations from the top layer, a 
partially passively stable omnidirectional gait can be achieved.

The keypoint gait generator is analytically computed as a function of the 
so-called \emph{gait phase} $\mu \in (-\pi,\pi]$, which is cyclically 
incremented over time based on the desired gait frequency. Every gait phase 
value uniquely corresponds to a particular instant of the stepping gait 
trajectory, where for example $\mu = 0$ corresponds to the begin of the double 
support phase transitioning from left support to right support, and $\mu = \pi$ 
corresponds to the nominal moment of foot strike of the left foot. The keypoint 
gait generator is computed in a target- and constraint-based manner, as opposed 
to being manually constructed, and explicitly considers that the robot may 
nominally walk at some pitch angle $p_{yN}$ relative to upright. This pitch 
angle defines a constant frame \fr{N} and corresponding plane N relative to the 
body frame \fr{B} of the robot, referred to as the \emph{nominal ground plane}, 
that during ideal nominal walking would indicate the planar level of the ground. 
Although entirely novel, the inner workings of the keypoint gait generator are 
beyond the scope of this paper.

\subsubsection{Higher level controller}
\seclabel{higher_level_controller}

The top layer of the presented gait architecture is generically referred to as 
the higher level controller, but more specifically in this case as the 
\emph{tilt phase controller}. Its task is to compute, based on sensory feedback, 
activations of the corrective actions that robustly stabilise the gait and 
ensure the robot remains balanced. The sources of sensory feedback can 
hypothetically include anything, but in this paper are strictly limited to a 
single 6-axis IMU.

\subsection{Corrective Actions}
\seclabel{corrective_actions}

Numerous corrective actions have been implemented in the gait, as shown in 
\figref{corrective_actions}. All cartesian actions are expressed in 
dimensionless form relative to the nominal ground plane N, in units of either 
the \emph{inverse leg scale} or \emph{tip leg scale} of the robot. These are 
respectively the vertical distances in the zero pose between the hip point and 
ankle point, and between the hip point and foot geometric centre. All 
rotation-based actions are expressed in the 2D tilt phase space, as pure tilt 
rotations relative to N. Pure \emph{tilt rotations} are rotations with a zero 
fused yaw component, i.e.\ pure rotations about an axis in the horizontal 
xy-plane. The implemented corrective action activations are as follows:
\begin{itemize}
\item The \textbf{arm tilt} $P_a = (p_{xa},p_{ya})$ to apply to the arms, to 
shift their CoMs and cause corresponding reaction moments.
\item The \textbf{support foot tilt} $P_s = (p_{xs},p_{ys})$ to apply to the 
feet during their respective support phases, with smooth transitions in the 
adjoining double support phases.
\item The \textbf{continuous foot tilt} $P_c = (p_{xc},p_{yc})$ to apply to the 
feet as constant offsets throughout the entire trajectory.
\item The dimensionless \textbf{hip shift} $\vect{s} = (s_x, s_y)$ to apply to 
the robot in units of inverse leg scale.
\item The \textbf{maximum hip height} $H_{max}$ to allow relative to the feet 
for the generated gait, in units of leg tip scale.
\item The \textbf{lean tilt} $P_l = (p_{xl},p_{yl})$ to apply to the robot 
torso, causing the robot to lean primarily via its hip joints.
\item The \textbf{swing out tilt} $P_o = (p_{xo},p_{yo})$ to smoothly apply to 
the midpoint of the swing trajectory, to adjust the path taken by the swing leg 
to its foot strike location.
\item The tilt $P_S\,{=}\,(p_{xS},p_{yS})$ defining the \textbf{swing ground 
plane} S relative to the N plane. This is the plane that is used to adjust the 
relative foot heights and tilts generated by the gait for orientation deviations 
in the trunk.
\item The \textbf{gait frequency} $f_g$ to use, in \si{\radian\per\second}, as 
the required instantaneous rate of change for updating the gait phase.
\end{itemize}
Note that all activations are expressed in a dimensionless manner so that almost 
exactly the same values can be used for robots of different scales. The 
configurable constants that are used throughout this gait also follow the same 
approach.

The corrective actions were not arbitrarily chosen, but were the result of an 
analysis of the conceivable strategies for balanced bipedal walking. A commanded 
gait velocity could easily be added to the list of corrective actions if an 
additional step size adjustment scheme is wished in parallel.

\section{Tilt Phase Controller}
\seclabel{tilt_phase_controller}

An overview of the feedback pipeline is shown in \figref{feedback_pipeline}. The 
individual actions are presented in detail in this section.

\subsection{Preliminaries}
\seclabel{preliminaries}

The tilt phase controller utilises a number of recurring filters and 
mathematical constructs, discussed as follows.

\subsubsection{Filters}
\seclabel{prelim_filters}

The mean and weighted line of best fit (WLBF) filters from \cite{Allgeuer2016a} 
have been taken and generalised to $n$ dimensions. The former computes the 
moving average of an $n$-dim vector, while the latter performs weighted 
time-based linear least squares regression to smooth and estimate the derivative 
of $n$-dim data. The advantages of WLBF filters over alternatives for the 
numerical computation of derivatives are discussed in depth in 
\cite{Allgeuer2016a}.

\subsubsection{Coerced interpolation}
\seclabel{prelim_coerced_interpolation}

Standard linear interpolation can lead to extrapolation outside of the interval 
domain. Coerced interpolation limits the input variable to ensure that the 
output cannot be outside the range of the two data points.

\subsubsection{Soft Coercion}
\seclabel{prelim_soft_coercion}

The soft coercion from \cite{Allgeuer2016a} has been used and extended 
ellipsoidally to $n$ dimensions. Given an input vector $\vect{x}$, the principal 
semi-axis lengths of the limiting ellipsoid $E$, and a soft coercion buffer $b$, 
symmetric soft coercion of buffer $b$ is applied radially along $\vect{x}$ to 
the radius of $E$. This is significantly better than applying soft limits along 
each axis independently, as the latter would result in unexpectedly large radial 
limits inbetween the principal axes.

\subsubsection{Smooth Deadband}
\seclabel{prelim_smooth_deadband}

The smooth deadband from \cite{Allgeuer2016a} has been used and extended 
ellipsoidally to $n$ dimensions. Given an input vector $\vect{x}$ and the 
principal semi-axis lengths of the deadband ellipsoid $E$, scalar smooth 
deadband is applied radially along $\vect{x}$ with a deadband radius of the 
radius of $E$.

\begin{figure}[!t]
\parbox{\linewidth}{\centering\includegraphics[width=1.00\linewidth]{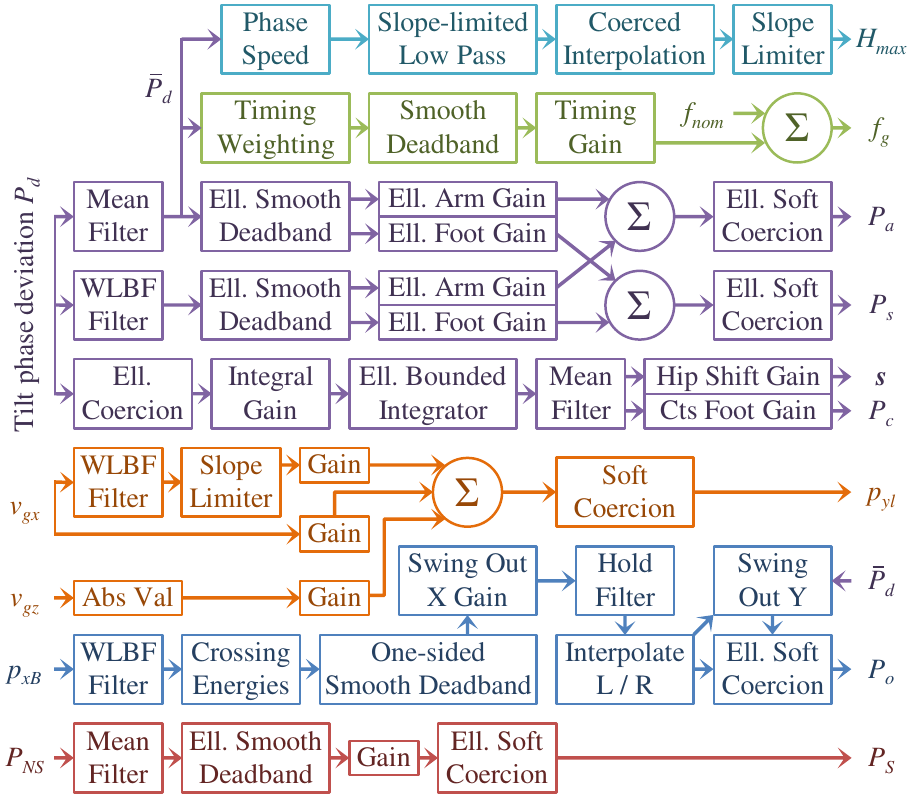}}
\caption{Overview of the tilt phase controller feedback pipeline.}
\figlabel{feedback_pipeline}
\vspace{-2ex}
\end{figure}
\subsection{Deviation Tilt}
\seclabel{deviation_tilt}

In the tilt phase controller, most of the activations of the corrective actions 
depend directly on how the robot is currently tilted relative to what is 
expected for the current gait phase. Based on the 3-axis accelerometer and 
gyroscope measurements from the IMU in the trunk of the robot, the tilt phase 
orientation $P_B = (p_{xB}, p_{yB})$ of the robot is estimated using a 3D nonlinear 
passive complementary filter \cite{Allgeuer2014}. Note that the yaw component is 
not included here as it does not directly contribute to the local balance state 
of the robot, and cannot be estimated drift-free based only on the IMU.

The two estimated tilt phase parameters follow an expected periodic pattern 
during ideal undisturbed walking of the robot, modelled as a map from $\mu \in 
(-\pi,\pi]$ to $\P^2$. Thus, by evaluation of this map, the expected tilt phase 
parameters $P_E = (p_{xE}, p_{yE})$ for the body of the robot can be calculated 
in every cycle. In this work, a sinusoidal map with an offset is used in both 
dimensions, as it sufficiently describes and generalises the observed data from 
the tested robots.

The corrective actions in the keypoint gait generator are defined so that they 
physically act to cause tilts relative to the N plane, just like the definitions 
of their respective activations (see \secref{corrective_actions}). Thus, an 
error term is constructed as a basis for feedback that quantifies the inverse of 
the unique tilt rotation relative to N that rotates the body frame \fr{B} to 
have the expected tilt phase parameters $P_E$. The so-called \emph{deviation 
tilt} $P_d = (p_{xd}, p_{yd})$ is calculated using
\begin{equation}
\begin{gathered}
q_d = q_y(p_{yN}) \, q_P(P_B)\conj \, q_z(\psi_E) \, q_P(P_E) \, q_y(-p_{yN}), \\
P_d = P_q(q_d\conj),
\end{gathered}
\eqnlabel{deviationtilt}
\end{equation}
where $\psi_E$ is the solution to $\yawof{q_d} = 0$, and $\yawof{\bigcdot}$ is 
the fused yaw of $\bigcdot\,$, $q_y(\theta)$ and $q_z(\theta)$ are the 
quaternions corresponding to pure y- and z-rotations by $\theta$, $p_{yN}$ is 
the nominal ground plane pitch, $q_P(\bigcdot)$ is the quaternion corresponding 
to the tilt phase $\bigcdot\,$, $\conj\mspace{-1mu}$ is the quaternion 
conjugation operator, and $P_q(\bigcdot)$ is the 2D tilt phase corresponding to
the quaternion $\bigcdot\,$.

As an illustratory guide, at this point if the negative of the calculated 
deviation tilt $P_d$ were for example to be applied as an activation to any of 
the rotation-based corrective actions, the effect would be to induce a tilt of 
the robot torso in some way towards its expected tilt phase parameters $P_E$. 
Thus, the feedback would act to return the robot towards nominal walking. Note 
however, that due to sensor noise and scaling issues, such direct feedback would 
not yet be effective.

\subsection{PD Feedback: Arm and Support Foot Tilt}
\seclabel{pd_feedback}

The most important thing for the stability of the robot on the short-term is to 
ensure that transient disturbances, such as for example pushes or steps on 
larger irregularities in the ground, are swiftly counteracted, and with little 
delay. 3D rotational proportional (P) and derivative (D) feedback components, 
activating the arm tilt and support foot tilt corrective actions, are used for 
this purpose. The arm tilt rotates the CoM of the arms out in the required 
direction relative to the N plane, so as to bias the CoM of the entire robot, 
and also to apply a reactive moment on the torso of the robot that helps 
mitigate the disturbance. At the same time, the support foot tilt applies smooth 
temporary corrections to the tilt of each foot during its respective single 
support phase, to push the robot back towards its expected orientation.

In order to reduce signal noise, the \textbf{P feedback} component first mean 
filters the deviation tilt using a small filter order to give $\bar{P}_d$. This 
tilt phase is then elliptically smooth deadbanded to ensure that P feedback only 
takes effect when the robot is non-negligibly away from its expected 
orientation. An elliptically directionally dependent gain is then applied to the 
resulting tilt phase, once independently for each the arm tilt and the support 
foot tilt, to get the corresponding P feedback components. The gain in each case 
is calculated from specifications of the required gains in the sagittal and 
lateral directions. Importantly, the directions of the final feedback components 
are both unchanged from $\bar{P}_d\mspace{1mu}$---all changes are purely radial.

For the \textbf{D feedback}, a smoothed derivative of $P_d$ is first computed 
using a 2D WLBF filter. A WLBF filter was chosen for its many advantages, 
including amongst other things for its favourable balance between robustness to 
high frequency noise and responsiveness to input transients 
\cite{Allgeuer2016a}. The computed derivative is elliptically smooth deadbanded 
to ensure that D feedback only takes effect if the robot torso has a 
non-negligible angular velocity relative to its expected orientation. Then, as 
for the P feedback, independent elliptically directionally dependent gains are 
applied to get the D feedback components for the two PD corrective actions.

Once the separate P and D components have been calculated, they are combined 
using tilt vector addition \cite{TiltPhase} and elliptically soft-coerced to 
obtain the final activations $P_a$ and $P_s$ (see \secref{corrective_actions}). 
Note that although it is not generally acceptable to just add 3D rotations, the 
special properties of the tilt phase space allow us to do just that in a 
meaningful, unambiguous, self-consistent and mathematically supported way. In 
fact, the tilt phase space turns tilt rotations into a well-defined vector space 
over $\R$, explaining why the scaling and addition of tilt phases that is used 
in this paper is actually mathematically valid and geometrically meaningful 
\cite{TiltPhase}. 

Tuning of the PD feedback is relatively straightforward, as there are only a few 
gains, and each gain has a clearly visible effect on the robot. The P feedback 
is tuned first, and then appropriated with D feedback to add damping to the 
system and limit oscillatory behaviour.

\subsection{I Feedback: Hip Shift and Continuous Foot Tilt}
\seclabel{i_feedback}

The implemented PD feedback works well for rejecting the majority of short-term 
transient disturbances, but if there are continued regular disturbances, or a 
systematic imbalance in the robot, the PD feedback in combination with other 
corrective actions will constantly need to act to oppose them. PD feedback can 
only act however, if there is a non-zero error present. Thus, without integral 
feedback on the hip shift and continuous foot tilt corrective actions, the 
system in such a case would at best settle with a steady state deviation to 
normal walking. The continuous foot tilt applies continuous tilt corrections to 
the generated orientations of the feet, while the hip shift applies an offset to 
the generated hip position relative to the feet. Both are applied relative to 
the N plane, and bias the balance of the robot in the desired direction to 
overcome systematic errors in the walking of the robot. The implemented I 
feedback can effectively reduce the need for fine tuning of the robot, and make 
the gait insensitive to small changes in the hardware or walking surface that 
would otherwise have been noticeable in the walking quality.

Starting with the deviation tilt $P_d$, standard elliptical coercion is first 
applied, the output of which is scaled by a scalar integral gain. A scalar gain 
is used instead of a directionally dependent one, so as not to distort the 
`aggregated' direction of instability once integration is applied. The initial 
coercion is useful to ensure that the integrated value is determined 
predominantly by small and consistent deviation tilts, rather than large and 
brief transients, which have little correlation to the finer balance of the 
robot. The coerced and scaled deviation tilt is passed to an \emph{elliptically 
bounded integrator}. This kind of integrator performs updates of 2D trapezoidal 
integration and elliptical soft coercion in each step. Note that the two steps 
are interlinked, as the output of the coercion is used as the starting point for 
the next integral update. Apart from providing the required integral behaviour 
to eliminate steady state errors and ensuring that the integral remains 
conveniently bounded, this special kind of integrator also inherently combats 
integral windup. The initial coercion of $P_d$ reduces the extent to which 
integrator windup is possible, but the elliptically bounded integrator ensures 
that the integral can move away from the elliptical boundary as quickly as it 
can get there, and cannot get stuck there due to over-integration. The dynamic 
response of the corrective actions is on a much quicker time scale than the 
integration, so this is the main type of windup concern.

The integrated tilt phase value is passed through a final mean filter to combat 
ripple, before being separately scaled to get the final corrective action 
activations $P_c$ and $\vect{s}$. The order of the mean filter is chosen to 
correspond exactly to the duration of an even number of steps at the nominal 
gait frequency. Due to the periodicity and general regularity of the gait, this 
leads to almost perfect cancellation of ripple. This would not be achievable 
with an IIR filter, which would also have the downside of not as efficiently 
forgetting old data.

During tuning, it is attempted to keep at least one of the elliptical integral 
bounds at 1. This form of normalisation makes the tuning of the integral and 
corrective action gains relatively simple and intuitive, as the former then 
inversely relates to the parameters of the initial elliptical coercion, and the 
latter then corresponds to the desired maximum magnitude of each respective 
corrective action.

\subsection{Leaning}
\seclabel{leaning}

Leaning at the hips could be activated based on I feedback, but this would 
promote suboptimal walking postures of the robot, in part because it directly 
changes the measured orientation of the trunk without this necessarily 
ameliorating the overall balance of the robot. Leaning by PD feedback would also 
be possible, but although maybe not immediately intuitively obvious, neither 
attempting to lean forwards nor backwards is particularly useful for dissipating 
energy when for example falling forwards. Pure hip rotations are only useful if 
they are performed quite significantly, early enough before tipping, and in 
specific controlled scenarios, e.g.\ clean push disturbances, purely sagittal 
direction, not walking or immediately stop after disturbance, and so on. In most 
other situations, reactive leaning has a negative impact on walking robustness. 
As such, only feedforward leaning components based on the gait command velocity 
are implemented. These seek to avoid disturbances due to changes in walking 
velocity before they even occur. The gait acceleration is first estimated using 
a WLBF filter followed by a slope limiter. A linear combination of the sagittal 
velocity $v_{gx}$, absolute turning velocity $\abs{v_{gz}}$ and sagittal gait 
acceleration is then taken and soft-coerced to give $p_{yl}$ (see 
\secref{corrective_actions}). This feedforward of sagittal leaning helps in 
particular during strong turns, and when starting and stopping forwards walking.

\subsection{Swing Out}
\seclabel{swing_out}

The robot is said to be on a lateral crossing trajectory if it has enough 
lateral momentum that it will tip over the outside of its support foot. This is 
a difficult situation, as no simple reactive stepping or waiting strategy can 
prevent the fall. Acting alongside the arm tilt and support foot tilt actions, 
the swing out tilt was specifically designed to allow recovery from crossing 
trajectories. When significant lateral energy is detected, the current swing leg 
is rotated outwards to bias the balance of the robot, and apply a reactive 
moment that dissipates crossing energy. The lateral tilt phase $p_{xB}$ is first 
smoothed and differentiated using a WLBF filter. The line of best fit is 
evaluated at the mean of the recorded data points so that the estimated phase 
$\tilde{p}_{xB}$ and phase velocity $\dot{\tilde{p}}_{xB}$ are synchronised in 
time. Modelling the behaviour of the lateral tilt phase as similar to a 
nonlinear pendulum gives that
\begin{equation}
G_X \bigl( \phi_X, \dot{\phi}_X \bigr) = \tfrac{1}{C^2} \dot{\phi}_X^{\,2} + 2(\cos{\phi_X} - 1), \eqnlabel{pendulumenergy}
\end{equation}
is constant over an undisturbed trajectory, where $X$ is $L$ or $R$ depending on 
the support foot, $C$ is a constant, and
\begin{align}
&\begin{aligned}
\phi_X &= \lambda(\tilde{p}_{xB} - p_{xX}), \\
\dot{\phi}_X &= \lambda \dot{\tilde{p}}_{xB},
\end{aligned} &
\mspace{15mu} \lambda &=
\begin{cases}
-1& \text{if $X$ is $L$,} \\
+1& \text{if $X$ is $R$,}
\end{cases}
\eqnlabel{crossingangledefn}
\end{align}
where $p_{xX}$ is the phase at the point of crossing over foot $X$. Adjusting 
the signs of the kinetic and potential energy components based on whether they 
are contributing to or hindering crossing, leads to the definition of 
\emph{crossing energy},
\begin{equation}
E_X = \tfrac{1}{C^2} \dot{\phi}_X^{\,2} \sgn(\dot{\phi}_X) + 2(\cos{\phi_X} - 1) \sgn(\phi_X). \eqnlabel{crossingenergy}
\end{equation}
$E_X$ is a $\mathcal{C}^1$ function of $\phi_X, \dot{\phi}_X$, is zero for 
lateral tilt phase trajectories that come to rest exactly on the verge of 
crossing, and is more positive the greater the severity of crossing. One-sided 
smooth deadband centred at some $E_{min}$ is then applied to the calculated 
$E_X$, i.e.\ $E_L, E_R$, values. The result is scaled to give an initial measure 
of how much swing out is required for each direction. The one-sided deadband 
ensures that the swing out is zero below an energy of $E_{min}$, and smoothly 
transitions to a linear relationship beyond that. A pair of hold filters is 
applied to ensure that the greatest activation over the most recent time is kept 
and used for each side. The filtered lateral swing out values are interpolated 
linearly based on the expected support conditions, via the gait phase. At this 
point, a sagittal swing out component is added that ensures that the resultant 
swing out is, within limits, in the direction of $\bar{P}_d$. The final 
resulting swing out tilt $P_o$ is then elliptically soft-coerced to ensure that 
the swing out stays within reasonable limits. The tuning of swing out is done by 
examining crossing trajectories of the robot. The $p_{xX}$ values are read from 
the points of inflection, and $C$ is chosen to give the most constant profiles 
possible of $G_X$. A suitable value for $E_{min}$ can then be guessed and 
trimmed by calculating what stationary value of $\phi_X$ it should correspond 
to.

\subsection{Swing Ground Plane}
\seclabel{swing_ground_plane}

While nominally the ground coincides with the N plane during walking, with 
disturbances this is no longer the case. This can cause premature foot strike, 
which is both destabilising and prevents the robot from taking the step size it 
should. The swing ground plane S defines the plane that is used to adjust the 
stepping trajectories to avoid such issues. This is different to most 
implementations of `virtual slope' \cite{Allgeuer2016a} \cite{Missura2015} in 
that it does not just linearly slant the foot motion profile---it analytically 
computes a smooth trajectory that respects the S plane at foot strike, yet 
intentionally presses into or eases off the ground immediately after, so as to 
actually apply a restoring moment to the robot. Standard virtual slope 
implementations can in fact decrease walking robustness, as the more the robot 
leans forwards for instance, the higher the feet are lifted at the front, and 
thus the less resistance there actually is to falling forwards.

The S plane is first computed by finding a pure tilt rotation relative to N that 
makes the N plane coincident with where the N plane would be if the robot had 
its expected orientation $P_E$. This pure tilt rotation is calculated using
\begin{equation}
P_{NS} = P_q \bigl( q_y(p_{yN}) \, q_P(P_B)\conj \, q_P(P_E) \, q_y(-p_{yN}) \bigr), \eqnlabel{swingplaneraw}
\end{equation}
where the same notation as in \eqnref{deviationtilt} is used. Note that $S 
\equiv N$ when the robot has its expected orientation, i.e.\ $P_B = P_E$. To 
reduce noise and prevent S plane adjustments from being made when walking is 
near nominal, a mean filter followed by elliptical smooth deadband is applied to 
$P_{NS}$. A nominally unit gain is then applied to allow the strength of the S 
plane feedback to be tuned if this helps with passive stability. The resulting 
tilt phase is then passed through elliptical soft coercion to ensure that it 
stays within limits. This yields the final activation $P_S$ of the swing plane 
corrective action.

\subsection{Timing}
\seclabel{timing}

Timing is an important feedback mechanism, in particular for other corrective 
actions like for example swing out to work most effectively. The same timing as 
used in \cite{Allgeuer2016a} has been implemented, only reformulated in terms of 
the lateral deviation tilt $p_{xd}$ instead of the fused roll deviation. This as 
output computes the required commanded gait frequency $f_g$ to use for updating 
the value of the gait phase in each cycle.

\subsection{Maximum Hip Height}
\seclabel{max_hip_height}

It can happen due to repeated disturbances or self-disturbances that the robot 
enters a semi-persistent limit cycle of oscillations, often sagittal. In such 
situations, limiting the height of the hips above the feet can help lower the 
CoM, and thereby increase the passive stability of the robot as greater 
rotations are then required for tipping. As such, by temporarily restricting the 
maximum hip height of the robot, unwanted oscillations of the robot can be 
dissipated.

\begin{figure}[!t]
\parbox{\linewidth}{\centering%
\includegraphics[width=\linewidth]{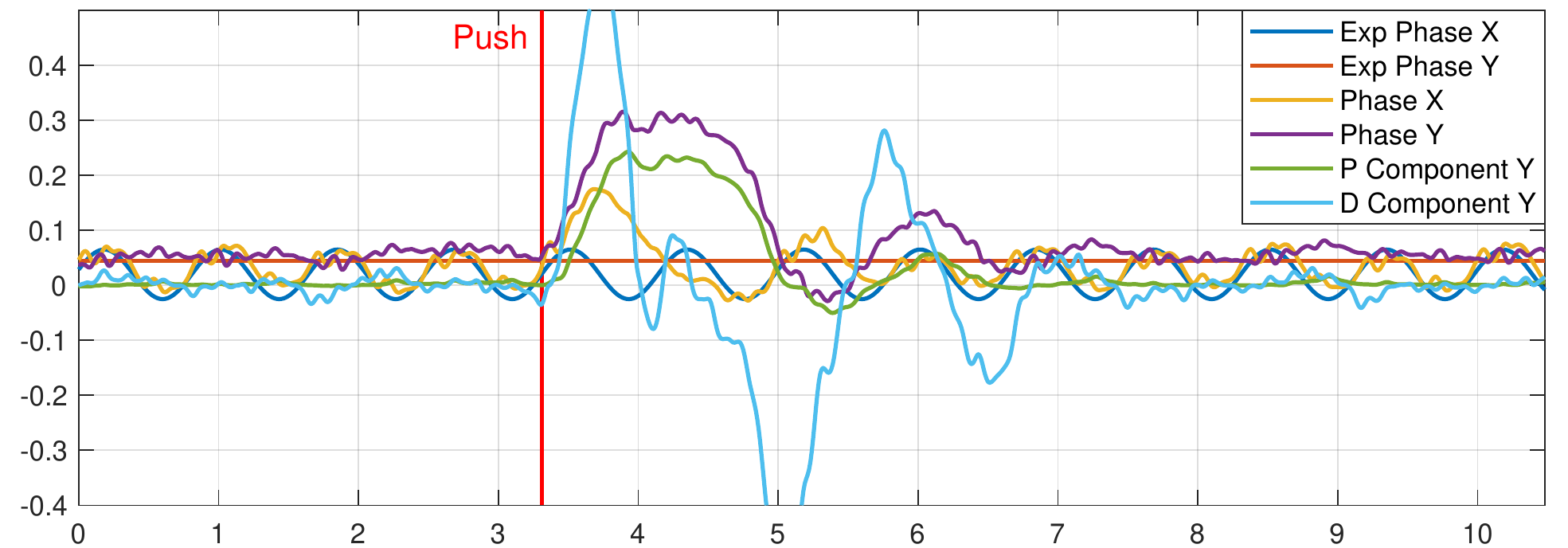}
\vspace{-3.7ex}\subcaption{The effect of the arm tilt and support foot tilt PD 
feedback in recovering balance after a diagonal push of significant 
destabilising power.}\figlabel{exp_pd}
\vspace{1.0ex}\includegraphics[width=\linewidth]{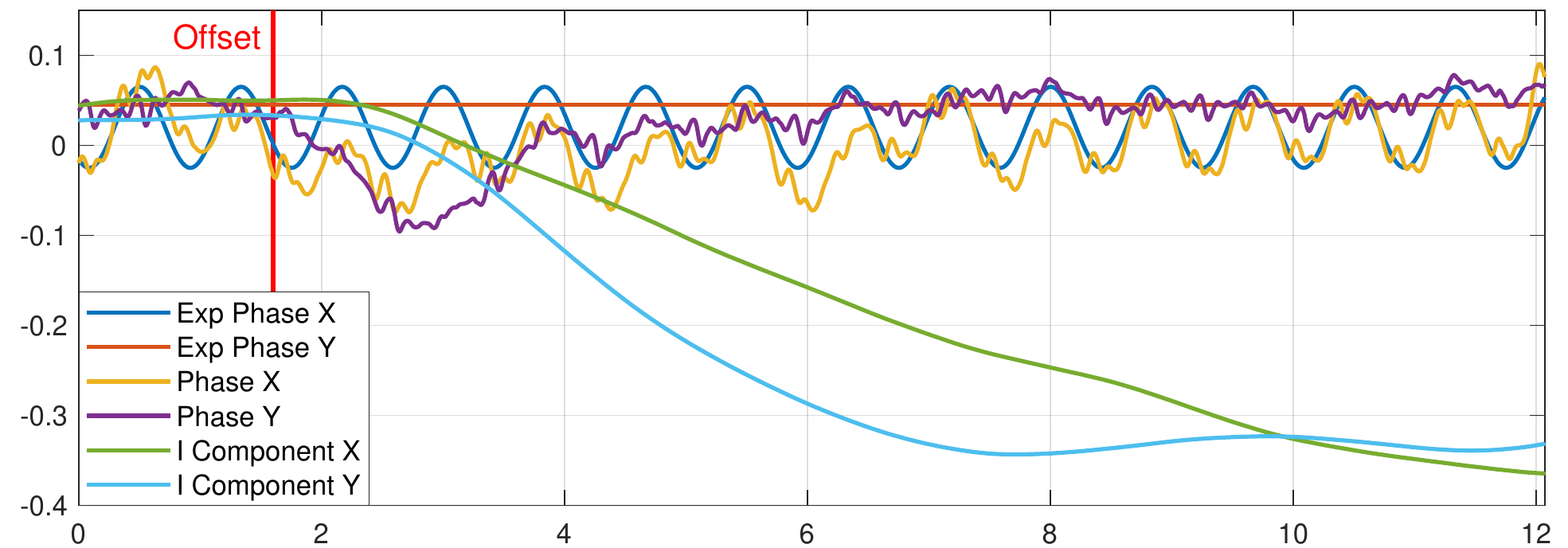}
\vspace{-3.7ex}\subcaption{The effect of the hip shift and continuous foot tilt 
I feedback after a sudden gait-unknown software offset is applied to the ankles 
of the robot.}\figlabel{exp_i}
\vspace{1.0ex}\includegraphics[width=\linewidth]{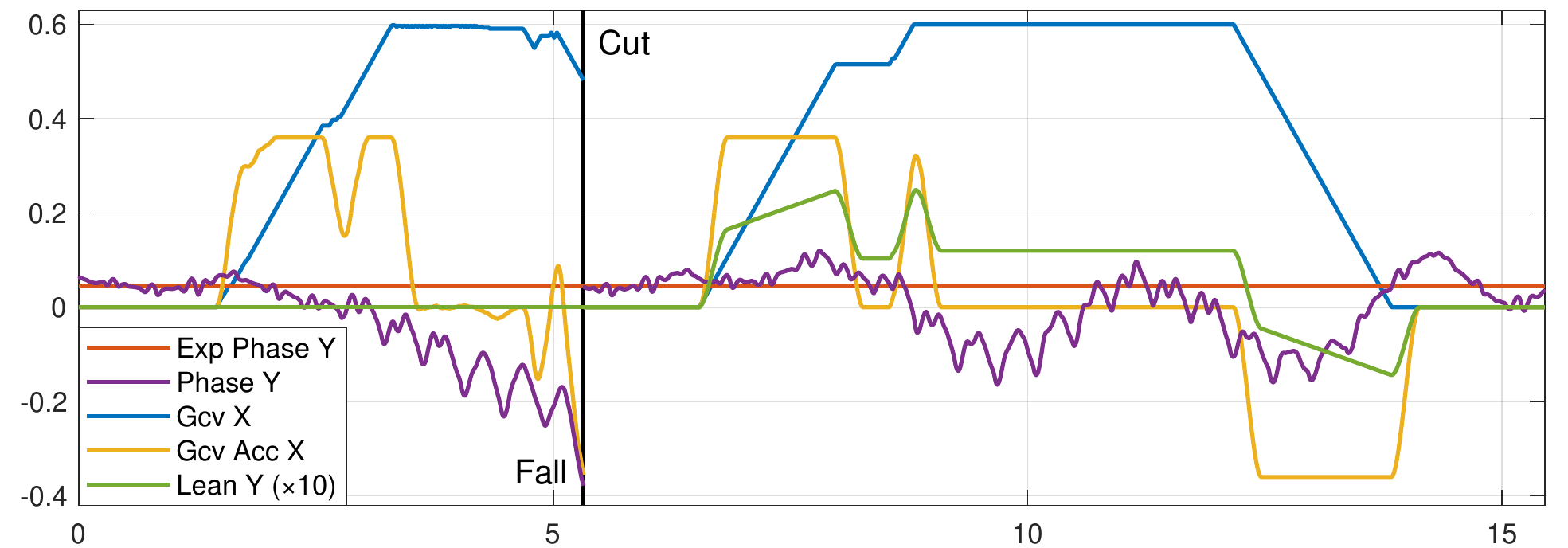}
\vspace{-3.7ex}\subcaption{The effect of leaning on sagittal walking. Without 
leaning (before cut), the forwards acceleration makes the robot tip over 
backwards even though walking on the spot is balanced. With leaning (after cut), 
there is no fall.}\figlabel{exp_leaning}
\vspace{1.0ex}\includegraphics[width=\linewidth]{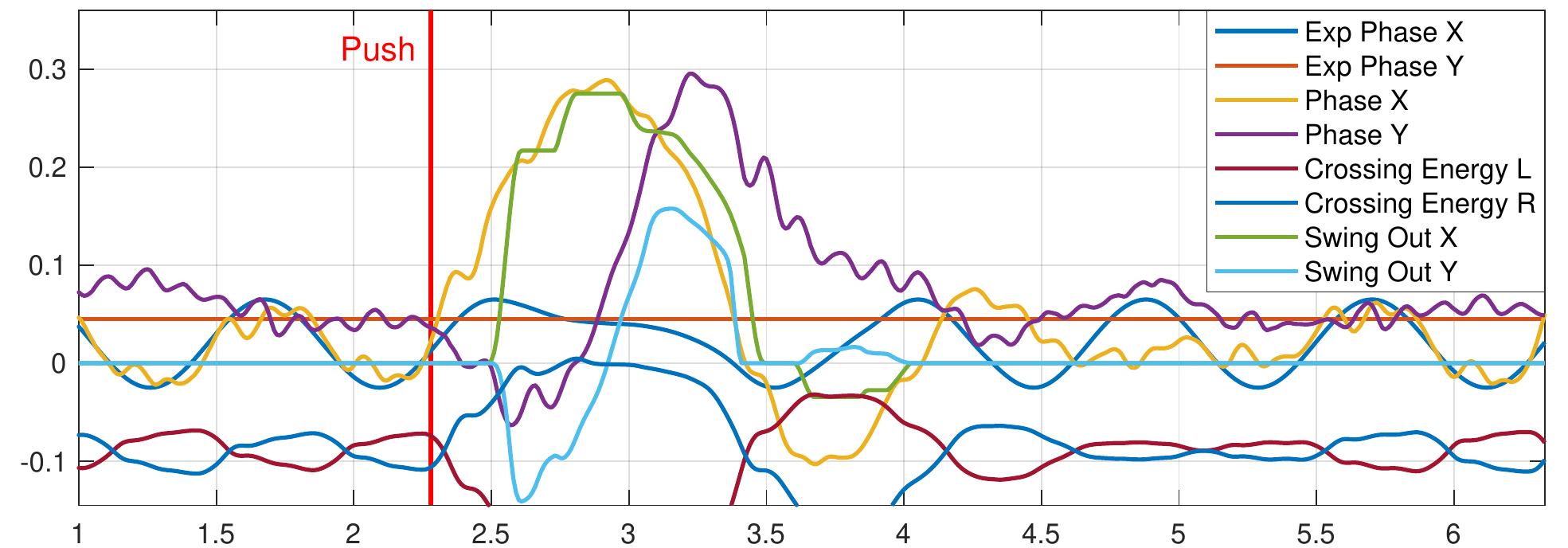}
\vspace{-3.7ex}\subcaption{The effect of the swing out tilt in recovering from a 
lateral push that would otherwise have led to a fall. The full 2D swing out 
allows the robot to remain balanced sagittally as well, in the \SI{1.2}{\second} 
it takes to return laterally.}\figlabel{exp_swingout}
}\caption{Plots of corrective actions acting on the real robot. The time axes 
are in units of \si{\second}, and the plotted quantities are by default in units 
of \si{\radian}. Exceptions are the D component (\si{\radian\per\second}), I 
components (\si{\radian\second}), `gcv' (dimensionless), and specific crossing 
energies (\si{\radian\squared}).}
\figlabel{exp_results_a}
\vspace{-2ex}
\end{figure}

A measure $I$ of the instability of the robot is first computed by applying a 
slope-limited low pass filter to normed speed values $s_d$ of the mean-filtered 
deviation tilt $\bar{P}_d$, i.e.\ to
\begin{equation}
s_d = \tfrac{1}{\Delta t} \norm{\Delta\bar{P}_d}. \eqnlabel{deviationtiltspeed}
\end{equation}
Note that only changes in orientation contribute to $I$. Note also that the low 
pass filter is nominally chosen to have a relatively long settling time, and 
that $\Delta\bar{P}_d$ can optionally be masked to only include sagittal 
components, if desired. Given the quantified instability $I$, coerced 
interpolation is used to map this to a desired range of maximum hip heights, so 
that greater levels of instability correspond to smaller allowed hip heights. A 
final slope limiter ensures that all changes to the resulting $H_{max}$ occur 
continuously, and suitably gradually.

The tuning of the hip height feedback essentially reduces to the choice of a 
settling time for the low pass filter, usually on the order of a few seconds, 
and the choice of an instability range to use for interpolation. The latter is 
tuned by artificially disturbing the robot and gauging as of what measured 
instability hip height feedback would be suitable.

\begin{figure}[!t]
\parbox{\linewidth}{\centering%
\vspace{1.0ex}\includegraphics[width=\linewidth]{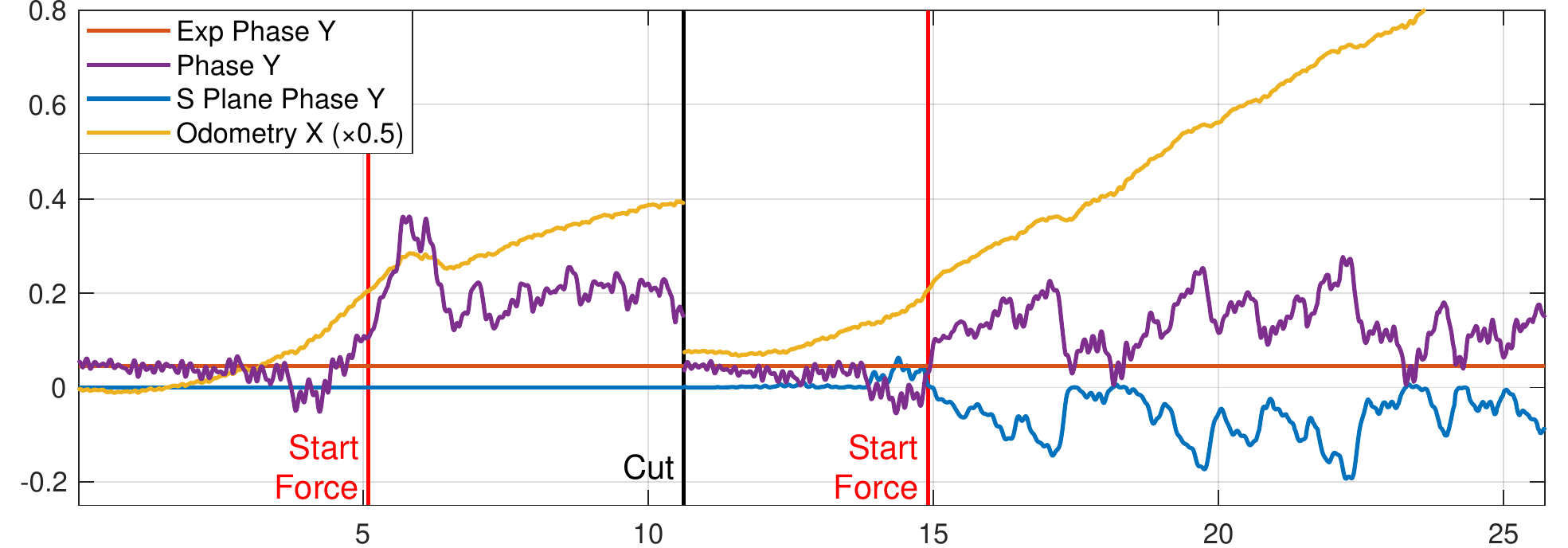}
\vspace{-3.7ex}\subcaption{Effect of the swing ground plane in reducing 
premature foot strike, demonstrated by applying a continuous forwards force to 
the robot that induces sagittal tilt. Before cut: S plane disabled, After cut: S 
plane enabled.}\figlabel{exp_planes}
\vspace{1.0ex}\includegraphics[width=\linewidth]{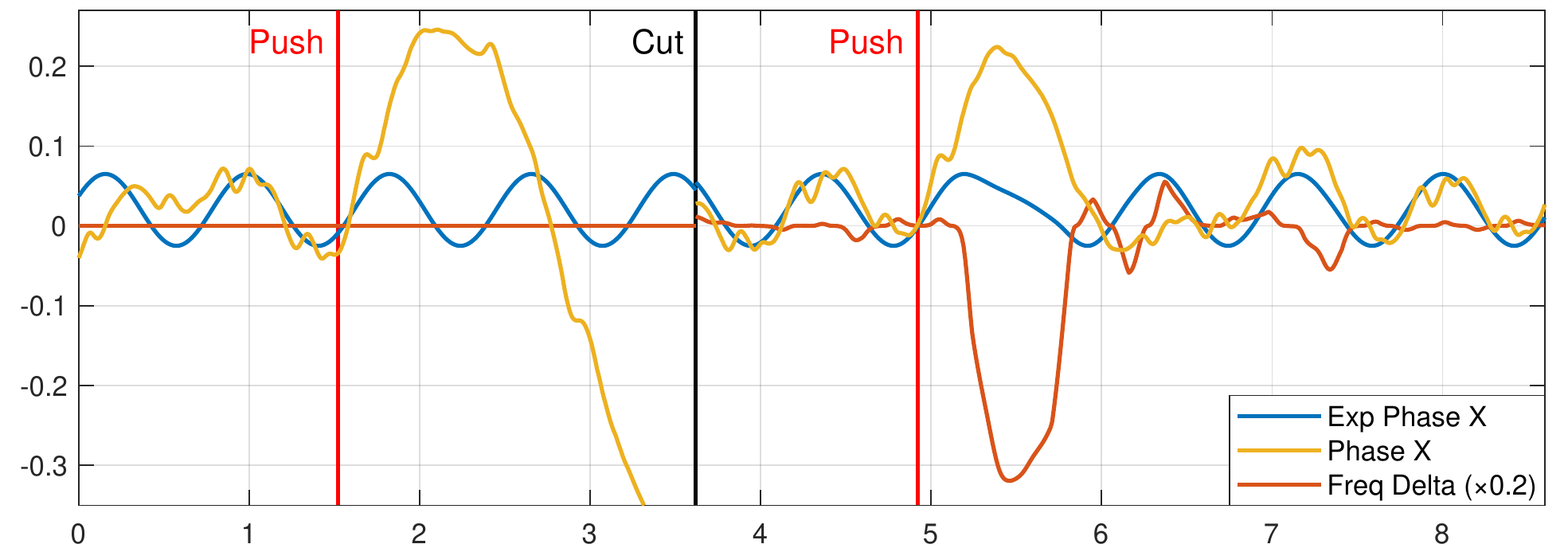}
\vspace{-3.7ex}\subcaption{The effect of timing feedback to avoid 
self-disturbances that can lead to a fall. Before cut: Timing disabled leads to 
fall, After cut: Timing enabled.}\figlabel{exp_timing}
\vspace{1.0ex}\includegraphics[width=\linewidth]{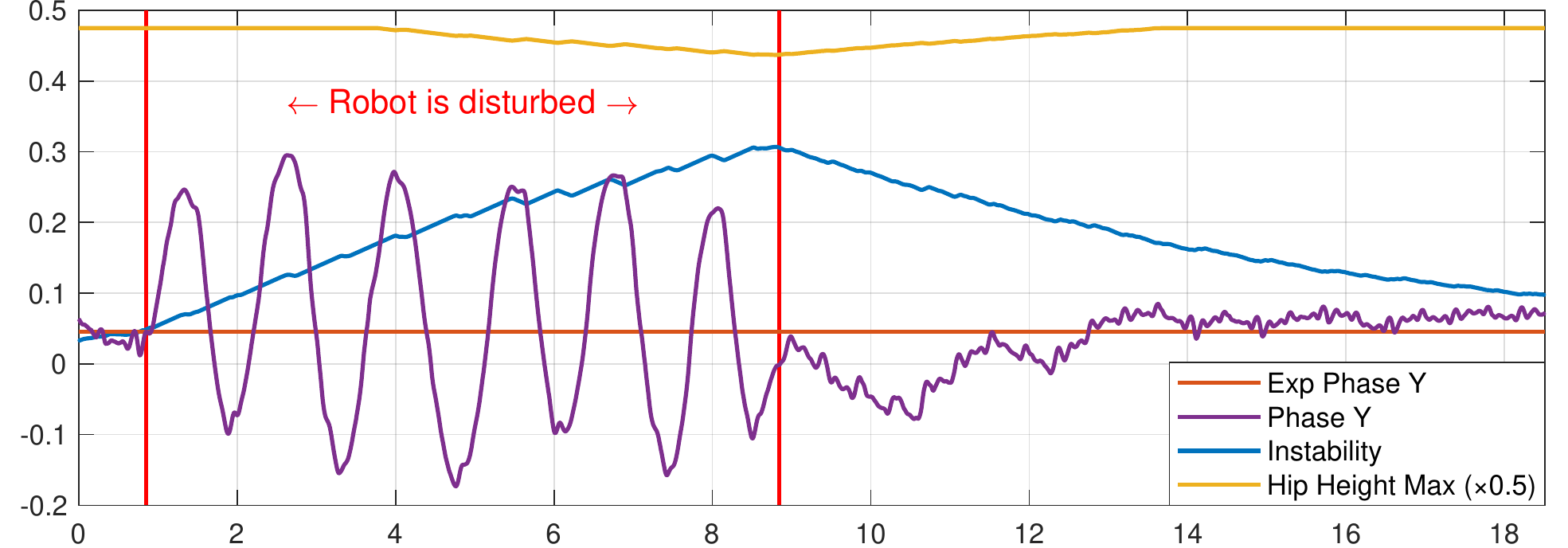}
\vspace{-3.7ex}\subcaption{Oscillations in the robot orientation, induced 
artificially here, cause the calculated instability to increase and limit the 
maximum allowed hip height.}\figlabel{exp_maxhipheight}
}\caption{Further plots of corrective actions acting on the real robot. The time 
axes are in units of \si{\second}, and the plotted quantities are by default in 
units of \si{\radian}. Exceptions are the odometry (\si{\metre}), frequency 
delta (\si{\hertz}), maximum hip height (dimensionless), and instability 
(\si{\radian\per\second}).}
\figlabel{exp_results_b}
\vspace{-2ex}
\end{figure}
\section{Experimental Results}
\seclabel{experimental_results}

The proposed feedback controller has been implemented in \cpp in the open-source 
\iguhop ROS software release \cite{iguhopSoftware}, which also supports the 
\noptwo robots. The entire controller takes just \SI{2.1}{\micro\second} to 
execute on a single \SI{3.5}{\giga\hertz} core. As such, it is expected that the 
implementation of this method at \SI{100}{\hertz} on even a modest 
microcontroller would be possible. Such portability is of great advantage in the 
area of low-cost robotics. Also, given the relative complexity of the gait and 
the diverse range of corrective actions, the number of important configuration 
constants has been kept rather low. The constants are in all cases also 
expressed in a way that they are dimensionless, easy to understand and tune, and 
more than often just the default values can be used due to these two factors.

\figref{exp_results_a} and \figref{exp_results_b} show plots of the tilt phase 
controller running on a real \iguhop. Experiments were performed to isolate and 
demonstrate the effect of the various corrective actions, in most cases with all 
other feedback turned off. The physical response of the robot to such 
disturbances can be seen in an adjoining 
video.\footnote{\scriptsize\url{https://youtu.be/spFqqktZ1s4}} In 
\figref{exp_pd}, it can be observed that the tilt phase corresponds closely to 
the expected waveforms, until a large diagonal push disturbs the robot. The PD 
activations quickly spike, preventing a forwards fall, and aiding the robot in 
returning to its expected tilt phase trajectory. Note that when the robot starts 
returning to upright, the sign of the D feedback changes to dampen the system 
and prevent excessive overshoot. In \figref{exp_i}, a gait-external offset to 
the ankle orientations is suddenly applied in software, with only I feedback 
enabled. Over time, 2D hip shifts and foot tilts are integrated up to negate the 
effect of the applied offset, and return the robot to nominal walking. In 
\figref{exp_leaning}, the robot is made to start walking forwards and then slow 
down and stop again shortly after. Without the sagittal leaning components the 
robot falls backwards, but with them the robot leans forwards initially and 
backwards at the end to add feedforward stability to the walking. In 
\figref{exp_swingout}, a large lateral push is applied that puts the robot on a 
crossing trajectory. Using the counterbalance of its free leg, energy is taken 
out of the swing, ensuring that the crossing energy does not surpass zero, which 
would indicate a non-returning trajectory. Note that during the long time spent 
on the support leg, y components of swing out are used to help prevent sagittal 
falls.

\begin{table}[!t]
\renewcommand{\arraystretch}{1.3}
\setlength\abovecaptionskip{0pt}
\caption{Number of Withstood Simulated Pushes of 20}
\tablabel{gazebo_push_results}
\centering
\begin{tabular}{r c c c c c c c c}
\hline
\textbf{Impulse (\si{\second\newton})} & 1.2 & 1.5 & 1.8 & 2.0 & 2.2 & 2.4 & 2.6 & 2.8\\
\hline
\textbf{Tilt phase} & 20 & 20 & 17 & 16 & 14 & 9 & 7 & 4\\
\textbf{Allgeuer \cite{Allgeuer2016a}} & 19 & 20 & 15 & 12 & 11 & 8 & 3 & 4\\
\hline
\end{tabular}
\vspace{-2ex}
\end{table}

In \figref{exp_planes}, a continuous forwards force is applied to the robot 
during forwards walking that causes the robot to tilt forwards. Without swing 
ground plane adjustment, the robot tends to walk `into the ground', and gets 
more stuck than if the swing ground plane is enabled. This is evidenced by the 
difference in sagittal gait odometry, also plotted. In \figref{exp_timing}, 
lateral pushes are applied to the robot that disrupt its natural walking rhythm. 
Without timing adjustments, this leads to a lateral fall. With adjustments 
however, the robot slows down its stepping motion when it detects the 
disturbance and tries to place its next foot when the lateral tilt is close to 
zero again. In \figref{exp_maxhipheight}, the robot is artificially disturbed 
over multiple seconds to demonstrate an extreme case of how oscillations lead to 
a higher quantified level of instability, and subsequently a reduction in hip 
height. In real walking, this is relevant for when the robot gets stuck in a 
limit cycle of oscillations, a situation that can occur, but is difficult to 
replicate intentionally. Lowering the hip height increases passive stability and 
changes the natural frequency of the dynamics of the tilting motions. Both 
factors often lead to damping of the oscillations.

The tilt phase controller was also evaluated quantitatively in simulation in 
Gazebo. Maximum forwards walking speed tests were performed over a 
\SI{4}{\metre} distance with a \SI{1}{\metre} run-up. A maximum mean velocity of 
\SI{45.7}{\centi\metre\per\second} was achieved, while with 
\cite{Allgeuer2016a}, \SI{30.5}{\centi\metre\per\second} was 
achieved.\footnote{\scriptsize\url{https://youtu.be/yY0kpUZpjO4}} Sets of 20 
pushes in a random direction were also simulated for a robot walking in place, 
with various different push 
strengths.\footnote{\scriptsize\url{https://youtu.be/6Zfo8Ndt8xQ}} The number of 
withstood pushes for each method is shown in \tabref{gazebo_push_results}. The 
proposed method achieves clearly better results than \cite{Allgeuer2016a}.

\section{Conclusion}
\seclabel{conclusion}

Walking does not always require overly complex stabilisation mechanisms to 
achieve high levels of robustness. In this paper, a feedback controller for 
robust bipedal walking has been presented that relies solely on measurements 
from a single 6-axis IMU in the trunk, and is applicable to low-cost robots with 
noisy sensors, imperfect actuation and limited computing power. No highly tuned 
or complex physical models are required, and great portability is ensured 
through the use of dimensionless parameters and configuration constants. The 
wide variety of corrective actions that are employed cover many different 
aspects of balanced walking, including both short-term and long-term stability. 
In summary, although conceptually simple, the tilt phase controller achieves 
genuinely good results.

\bibliographystyle{IEEEtran}
\bibliography{IEEEabrv,ms}

\begin{thebibliography}{10}
\providecommand{\url}[1]{#1}
\csname url@samestyle\endcsname
\providecommand{\newblock}{\relax}
\providecommand{\bibinfo}[2]{#2}
\providecommand{\BIBentrySTDinterwordspacing}{\spaceskip=0pt\relax}
\providecommand{\BIBentryALTinterwordstretchfactor}{4}
\providecommand{\BIBentryALTinterwordspacing}{\spaceskip=\fontdimen2\font plus
\BIBentryALTinterwordstretchfactor\fontdimen3\font minus
  \fontdimen4\font\relax}
\providecommand{\BIBforeignlanguage}[2]{{%
\expandafter\ifx\csname l@#1\endcsname\relax
\typeout{** WARNING: IEEEtran.bst: No hyphenation pattern has been}%
\typeout{** loaded for the language `#1'. Using the pattern for}%
\typeout{** the default language instead.}%
\else
\language=\csname l@#1\endcsname
\fi
#2}}
\providecommand{\BIBdecl}{\relax}
\BIBdecl

\bibitem{Kryczka2015}
P.~Kryczka, P.~Kormushev, N.~Tsagarakis, and D.~Caldwell, ``Online regeneration
  of bipedal walking gait optimizing footstep placement and timing,'' in
  \emph{Int. Conf. on Intell. Robots and Systems (IROS)}, 2015.

\bibitem{Allgeuer2016a}
P.~Allgeuer and S.~Behnke, ``Omnidirectional bipedal walking with direct fused
  angle feedback mechanisms,'' in \emph{Proceedings of 16th Int. Conf. on
  Humanoid Robots (Humanoids)}, Canc\'un, Mexico, 2016.

\bibitem{iguhopSoftware}
\BIBentryALTinterwordspacing
{NimbRo}. (2018, Jul) {igus Humanoid Open Platform ROS Software}. [Online].
  Available: \url{https://github.com/AIS-Bonn/humanoid_op_ros}
\BIBentrySTDinterwordspacing

\bibitem{Wieber2006}
P.-B. Wieber, ``Trajectory free linear model predictive control for stable
  walking in the presence of strong perturbations,'' in \emph{IEEE Int. Conf.
  on Humanoid Robots (Humanoids)}, Genova, Italy, 2006.

\bibitem{Kajita2003}
S.~Kajita, F.~Kanehiro, K.~Kaneko, K.~Fujiwara, K.~Harada, K.~Yokoi, and
  H.~Hirukawa, ``Biped walking pattern generation by using preview control of
  zero-moment point,'' in \emph{Int. Conf. on Rob. and Auto.}, 2003.

\bibitem{Urata2011}
J.~Urata, K.~Nshiwaki, Y.~Nakanishi, K.~Okada, S.~Kagami, and M.~Inaba,
  ``Online decision of foot placement using singular {LQ} preview regulation,''
  in \emph{IEEE Int. Conf. on Humanoid Robots}, 2011.

\bibitem{Feng2013}
S.~Feng, X.~Xinjilefu, W.~Huang, and C.~Atkeson, ``{3D} walking based on online
  optimization,'' in \emph{Proceedings of 13th IEEE-RAS Int. Conference on
  Humanoid Robots (Humanoids)}, Atlanta, USA, 2013.

\bibitem{Tedrake2015}
R.~Tedrake, S.~Kuindersma, R.~Deits, and K.~Miura, ``A closed-form solution for
  real-time {ZMP} gait generation and feedback stabilization,'' in \emph{Int.
  Conf. on Humanoid Robots (Humanoids)}, Seoul, Korea, 2015.

\bibitem{Kajita2017}
S.~Kajita, M.~Benallegue, R.~Cisneros, T.~Sakaguchi, S.~Nakaoka, M.~Morisawa,
  K.~Kaneko, and F.~Kanehiro, ``Biped walking pattern generation based on
  spatially quantized dynamics,'' in \emph{Proceedings of 17th Int. Conf. on
  Hum. Robots (Humanoids)}, Birmingham, UK, 2017.

\bibitem{TiltPhase}
P.~Allgeuer and S.~Behnke, ``Tilt rotations and the tilt phase space,'' in
  \emph{Proceedings of 18th Int. Conf. on Humanoid Robots (Humanoids)},
  Beijing, China, 2018.

\bibitem{Allgeuer2015}
P.~Allgeuer and S.~Behnke, ``{F}used {A}ngles: {A} representation of body
  orientation for balance,'' in \emph{IROS Conf.}, Hamburg, Germany, 2015.

\bibitem{Allgeuer2018a}
P.~Allgeuer and S.~Behnke, ``Fused angles and the deficiencies of {E}uler
  angles,'' in \emph{Int. Conf. on Intell. Robots and Systems (IROS)}, 2018.

\bibitem{Allgeuer2014}
P.~Allgeuer and S.~Behnke, ``Robust sensor fusion for biped robot attitude
  estimation,'' in \emph{Proceedings of 14th IEEE-RAS Int. Conference on
  Humanoid Robotics (Humanoids)}, Madrid, Spain, 2014.

\bibitem{Missura2015}
M.~Missura, ``Analytic and learned footstep control for robust bipedal
  walking,'' Ph.D. dissertation, University of Bonn, 2015.

\end{thebibliography}

\end{document}